\documentclass[10pt,journal,compsoc]{IEEEtran}

\usepackage[nocompress]{cite}
 \usepackage[caption=false,font=footnotesize,labelfont=sf,textfont=sf]{subfig}

\let\MYorigsubfloat\subfloat
\renewcommand{\subfloat}[2][\relax]{\MYorigsubfloat[]{#2}}

\ifCLASSOPTIONcompsoc
  \usepackage[nocompress]{cite}
\else
  \usepackage{cite}
\fi
\ifCLASSINFOpdf
\else
\fi
\ifCLASSOPTIONcompsoc
 \usepackage[caption=false,font=footnotesize,labelfont=sf,textfont=sf]{subfig}
\else
 \usepackage[caption=false,font=footnotesize]{subfig}
\fi
\usepackage{graphicx,pstricks}
\usepackage{graphics}
\usepackage{moreverb}
\usepackage{palatino}
\usepackage{indentfirst}
\usepackage{xcolor}
\usepackage{svg}
\usepackage{textcomp}
\usepackage{amssymb}
\usepackage[version=4]{mhchem}
\usepackage{siunitx}
\usepackage{longtable,tabularx}
\usepackage{verbatim}
\usepackage{docmute}

\usepackage[ruled,vlined,linesnumbered]{algorithm2e}

\begin{document}
\title{Camera Calibration from a Single Imaged Ellipsoid: A Moon Calibration Algorithm}

\author{Kalani R. Danas Rivera and Mason A. Peck

\IEEEcompsocitemizethanks{\IEEEcompsocthanksitem K. Danas Rivera is a PhD Candidate with the Department
of Mechanical and Aerospace Engineering, Cornell University, Ithaca,
NY, 14850.\protect\\
E-mail: krd76@cornell.edu
\IEEEcompsocthanksitem M. A. Peck is an associate professor in the Department of Mechanical and
Aerospace Engineering with the Cornell University, Ithaca, NY.}
}


\IEEEtitleabstractindextext{%
\begin{abstract}
This work introduces a method that applies images of the extended bodies in the solar system to spacecraft camera calibration. The extended bodies consist of planets and moons that are well-modeled by triaxial ellipsoids. When imaged, the triaxial ellipsoid projects to a conic section which is generally an ellipse. This work combines the imaged ellipse with information on the observer's target-relative state to achieve camera calibration from a single imaged ellipsoid. As such, this work is the first to accomplish camera calibration from a single, non-spherical imaged ellipsoid. The camera calibration algorithm is applied to synthetic images of ellipsoids as well as planetary images of Saturn's moons as captured by the \textit{Cassini} spacecraft. From a single image, the algorithm estimates the focal length and principal point of \textit{Cassini's} Narrow Angle Camera within 1.0 mm and 10 pixels, respectively. With multiple images, the one standard deviation uncertainty in focal length and principal point estimates reduce to 0.5 mm and 3.1 pixels, respectively. Though created for spacecraft camera calibration in mind, this work also generalizes to terrestrial camera calibration using any number of imaged ellipsoids.

\end{abstract}

}

\maketitle

\IEEEdisplaynontitleabstractindextext
\IEEEpeerreviewmaketitle

\section{Introduction}
\subsection{Camera Calibration}
\IEEEPARstart{O}{ptical} Navigation (OPNAV) establishes estimates of a spacecraft's relative state using a calibrated onboard camera. Inaccuracies in the camera's intrinsic parameters propagate to inaccurate OPNAV estimates. Accurate knowledge of the camera's parameters requires accurate and precise camera calibration procedures. Full camera calibration estimates a variety of camera parameters \cite{zisserman2004multiple}, however we reserve the discussion to a camera's intrinsic, geometric calibration parameters. Geometric calibration solves for a camera's geometric distortion due to optics, effective focal length, and alignment. The effective focal length establishes the distance from the camera center to the focal or image plane~\cite{zisserman2004multiple}. The alignment addresses the location of the boresight's intersection with the focal plane (i.e., principal point). For spacecraft cameras, ground-based calibration provides preliminary values for a camera's intrinsic parameters. A second calibration occurs in space and images a star cluster such as the Pleiades for more precise and accurate estimates of the intrinsic camera parameters \cite{westCassiniISS,porco2004cassini,knowles2016cassini}.

\subsection{Calibration of a Camera's Intrinsic Parameters from a Planet's Horizon}
The majority of the solar system planets and moons resemble ellipsoids. As ellipsoids, OPNAV work in horizon-based navigation \cite{christian2021tutorial,christian2012onboard,christian2015optical} estimates a spacecraft's relative state using a calibrated camera. Given horizon-based navigation is a solved problem, we express interest in the inverse problem of estimating a camera's parameters given a spacecraft's relative state. In other words, is camera calibration possible using a planet's horizon? 

Camera calibration methods using ellipsoids and ellipses as imaging targets exist in the calibration literature ~\cite{yang2000planar,sun2015accurate,huang2015common,yang2020camera,su2020novel,hu2001camera}. However for the application of planet horizons, the minimum imaging target requirement and model assumptions in the existing literature are problematic. The calibration algorithms in Refs.~\cite{yang2000planar,sun2015accurate,huang2015common,yang2020camera,su2020novel,hu2001camera} require a minimum of $N\geq 2$ imaging ellipsoids or ellipses. Though imaging two bodies in the same image is possible 
, it requires favorable geometry and considerable planning to occur.
Imaging a single planet represents the general case experienced by most spacecraft. Additionally, camera calibration algorithms in Refs.~\cite{yang2000planar,sun2015accurate,huang2015common,yang2020camera,su2020novel,hu2001camera} specifically address images of spherical or circular imaging targets. As ellipsoidal bodies, planets are generally not spheres, and their apparent horizon projects to a conic section which is generally an ellipse and not a circle. As is, the existing methods only apply to the specific case of a nadir-pointed (i.e. pointing at planet center) spacecraft imaging a spherical planet, and a model mismatch exists for all other cases.

Camera calibration from a planet's horizon requires an algorithm that takes the single image of any arbitrary ellipsoid as input. We propose an algorithm that requires $N=1$ ellipsoidal targets for intrinsic camera calibration from a single image.

\section{Proposed Method}
Onboard spacecraft cameras are an immense sensing asset for onboard optical navigation. OPNAV methods enable spacecraft to fulfill their stringent navigation requirements and all rely on a calibrated camera for successful integration. For the case of spacecraft implementing OPNAV near an ellipsoidal planetary body, this work considers using the planetary body's horizon conic to calibrate the spacecraft camera's intrinsic parameters. 

\section{Conic-to-Conic Mapping}
\subsection{Reference Conic $\mathcal{C}$}
\begin{figure}[tbp]
    \centering
    \includegraphics[width=0.3\textwidth]{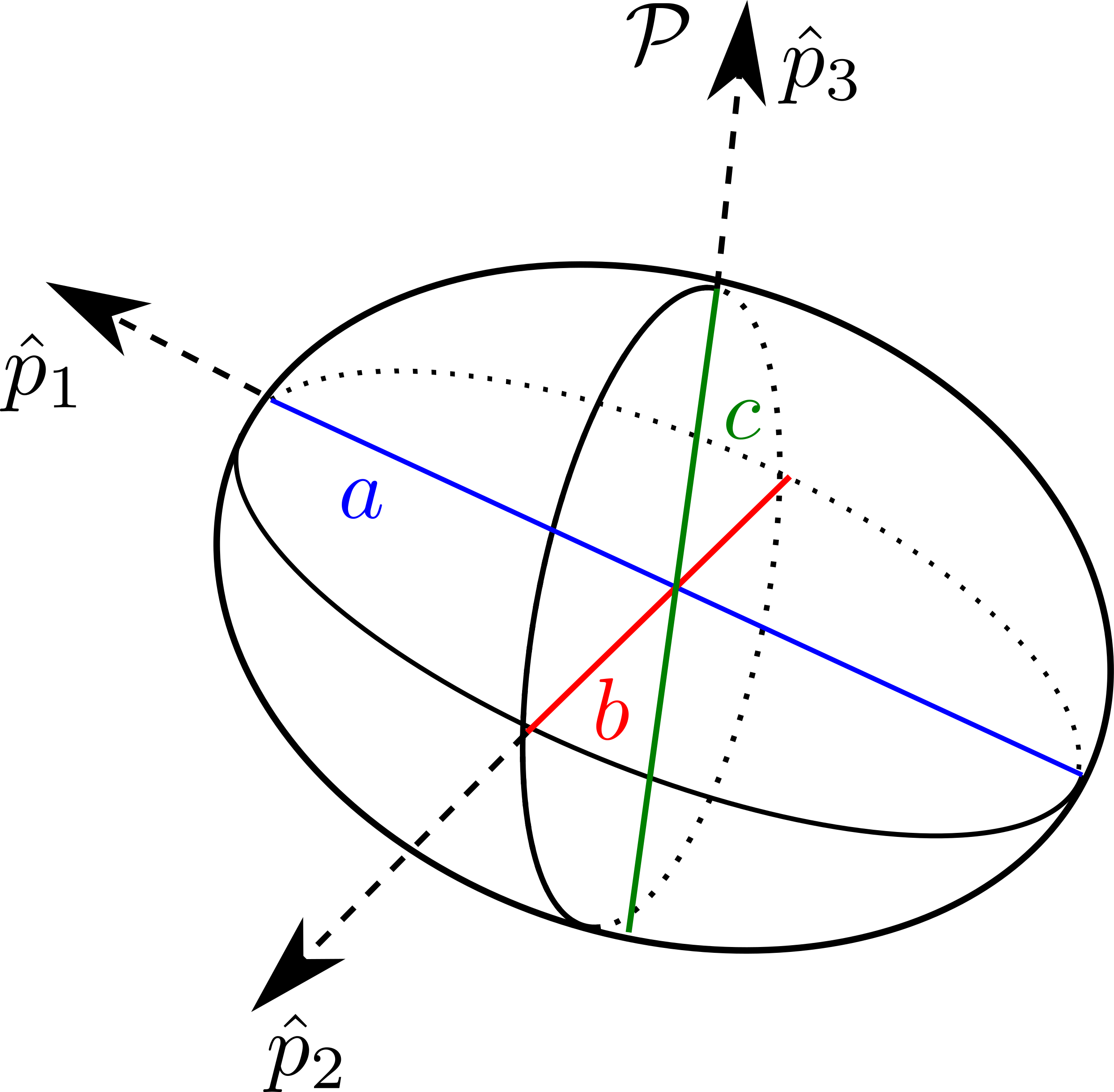}
    \caption{Triaxial Ellipsoid with Principal Axes ($\hat{p}_1, \hat{p}_2, \hat{p}_3 $)}
    \label{fig:triaxial_ellipsoid}
\end{figure}

Extended solar system bodies such as planets and moons resemble ellipsoids. For an ellipsoid, points on its surface $p_P\triangleq \begin{bmatrix} p_x & p_y & p_z\end{bmatrix}^T$ satisfy the following
\begin{equation} \label{calib_eq:quadric_surface}
    \begin{bmatrix}
     p_x \\ p_y \\ p_z
    \end{bmatrix}^T
     \begin{bmatrix}
      1/a^2 & 0 & 0 \\
      0 & 1/b^2 & 0  \\
      0 & 0 & 1/c^2
     \end{bmatrix}
    \begin{bmatrix}
     p_x \\ p_y \\ p_z
    \end{bmatrix}
    = p_P^T \mathcal{A}_P p_P
    = 1 
\end{equation}
equation for a quadric surface. The matrix
\begin{equation}
     \mathcal{A}_P \triangleq
     \begin{bmatrix}
      1/a^2 & 0 & 0 \\
      0 & 1/b^2 & 0  \\
      0 & 0 & 1/c^2
     \end{bmatrix}
 \end{equation}
defines the shape matrix of an ellipsoid in terms of its semi-major radii $a,b,$ and $c$ about its principal axes as shown in Fig.~\ref{fig:triaxial_ellipsoid}. Subscript $P$ denotes the vector/matrix in the planet's principal coordinate system.
National Aeronautics and Space Administration's (NASA) Navigation and Ancillary Information Facility (NAIF) maintains SPICE kernels that provide a host of parameters for planetary bodies including the $a,b,$ and $c$ of their respective best-fit ellipsoid as well as their position and orientation at a given epoch~\cite{acton1996ancillary,acton2018look}. When building $\mathcal{A}_P$ we refer to SPICE kernels for the appropriate $a,b,$ and $c$ values.

Shape matrix $\mathcal{A}_P$ influences the ellipsoid's perspective projection as imaged by an observer at planet-relative position $r_P$. From the observer, surface point $p_P$ exists along the line-of-sight (LOS) vector $\hat{e}_P$ at some distance $\lambda$. From vector addition,
\begin{equation} \label{calib_eq:p_redefined}
    p_P = r_P + \lambda \hat{e}_P
\end{equation}
provides $p_P$ in terms of $r_P$, $\hat{e}_P$, and $\lambda$. Substituting Eq.~\eqref{calib_eq:p_redefined} into Eq.~\eqref{calib_eq:quadric_surface} and re-arranging provides
\begin{equation}
    \hat{e}^T \mathcal{A}_{p} \hat{e} \lambda^2 + 2 r_P^T \mathcal{A}_{p} \hat{e} \lambda + (r_P^T \mathcal{A}_{p} r_P - 1) = 0
\end{equation}
a quadratic expression in terms of unknown $\lambda$. The familiar quadratic formula solves for unknown $\lambda$
\begin{equation}
\begin{aligned}
    \lambda_1, \lambda_2 = \frac{-2 r_P^T \mathcal{A}_{p} \hat{e}_P}{2 \hat{e}_P^T \mathcal{A}_{p} \hat{e}_P} \pm \\    
     \frac{ \sqrt{ 4 \hat{e}_P^T (\mathcal{A}_{p} r_P r_P^T \mathcal{A}_{p} ) \hat{e}_P - 4 \hat{e}_P^T ( (r_P^T \mathcal{A}_{p} r_P - 1) \mathcal{A}_{p})\hat{e}_P } }{2 \hat{e}_P^T \mathcal{A}_{p} \hat{e}_P}.
\end{aligned}
\end{equation}
As a quadratic expression, two roots exist for $\lambda$. However as Refs.~\cite{christian2012onboard,christian2015optical,christian2021tutorial} point out, $p_P$ lying tangent to the ellipsoid have repeated roots such that the discriminant becomes zero. The locus of $p_P$ tangent to the ellipsoid  describes the apparent outline or horizon of the ellipsoid as viewed by the observer at $r_P$. For horizon points, setting the discriminant to zero reduces to
\begin{equation}
    \hat{e_P}^T \left( \mathcal{A}_{p} r_P r_P^T \mathcal{A}_{p} - (r_P^T \mathcal{A}_{p} r_P - 1)\mathcal{A}_{p}  \right)\hat{e}_P
\end{equation}
for which the inner matrix
\begin{equation} \label{calib_eq:cone_world}
    \mathcal{C}_{p} \triangleq \mathcal{A}_{p} r_P r_P^T \mathcal{A}_{p} - (r_P^T \mathcal{A}_{p} r_P - 1) \mathcal{A}_{p}
\end{equation}
defines the conic formed by the horizon under perspective projection. The conic may be a circle, hyperbola, or parabola, but the horizon's projection is generally an ellipse. In this work we prefer expressing $\mathcal{C}_P$ in camera coordinates as given by
\begin{equation}
    \mathcal{C}_C \triangleq T_P^C \mathcal{C}_P T_C^P
\end{equation}
where $T_P^C$ is the coordinate transformation matrix from planet-fixed principal coordinates ($P$) to spacecraft camera coordinates ($C$).

\begin{figure}[htbp!]
    \centering
    \includegraphics[width=0.4\textwidth]{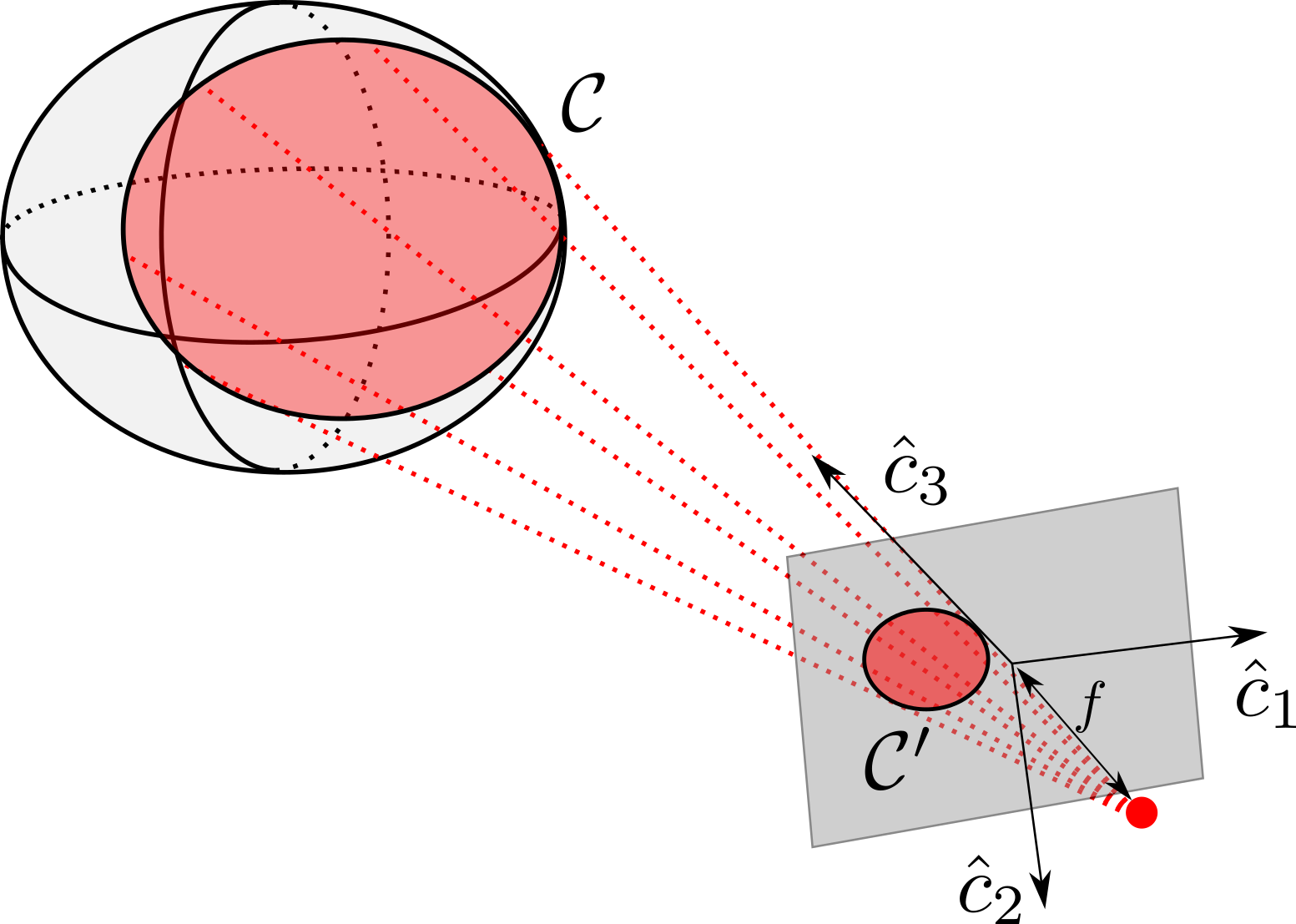}
    \caption{Perspective Projection of Ellipsoid}
\end{figure}

\subsection{Imaged Conic $\mathcal{C'}$}
From Eq.~\eqref{calib_eq:cone_world}, horizon points project to the observer as a conic. An imager observes the horizon projection on its image sensor in pixel coordinates $(u,v)$. Recovering the imaged horizon conic requires fitting all imaged horizon points to a conic section.  Conic-fitting based on the algebraic distance metric as in
\begin{equation} \label{calib_eq:conic_algebraic}
    Au^2 + Buv + Cv^2 + Du + Ev + F = 0
\end{equation}
provides coefficients $A,B,C,D,E,$ and $F$ of the best conic-fit. A plethora of conic-fitting algorithms exist, but we apply Ref.~\cite{kanatani2011hyper}'s method in this work for a direct, unbiased fit. Expressing Eq.~\eqref{calib_eq:conic_algebraic} into compact notation through homogeneous pixel coordinates $\overline{u}=\begin{bmatrix} u & v & 1 \end{bmatrix}^T$ reduces to 
\begin{equation}
    \begin{bmatrix}
     u \\ v \\ 1
    \end{bmatrix}^T
    \begin{bmatrix}
     A & B/2 & D/2 \\
     B/2 & C & E/2 \\
     D/2 & E/2 & F \\
    \end{bmatrix}
    \begin{bmatrix}
     u \\ v \\ 1
    \end{bmatrix}
    = \overline{u}^T \mathcal{C}' \overline{u}
    = 0 
\end{equation}
where
\begin{equation}
    \mathcal{C'} \triangleq
    \begin{bmatrix}
     A & B/2 & D/2 \\
     B/2 & C & E/2 \\
     D/2 & E/2 & F \\
    \end{bmatrix}
\end{equation}
denotes the imaged horizon conic in terms of the fitted coefficients.

\subsection{Action of Calibrated Camera}
An observer's relative position $r_P$ and attitude $T_P^C$ dictate the apparent horizon conic. The image sensor captures the imaged apparent horizon conic in $(u,v)$ coordinates. The reference and imaged conics relate to each other by the camera's calibration matrix $K$. The proportionality
\begin{equation} \label{calib_eq:proportionality}
    K^T \mathcal{C'} K \propto \mathcal{C}
\end{equation}
describes the conic-to-conic mapping between reference and imaged conics. Both sides of Eq.~\eqref{calib_eq:proportionality} are equivalent up to a scale factor. The camera calibration matrix relates an image from focal plane normalized coordinates to pixel coordinates as in
\begin{equation}
    \begin{bmatrix} u \\ v \\ 1 \end{bmatrix} = 
    \begin{bmatrix}
     f/\mu_x & \gamma & u_o \\
     0 & f/\mu_y & v_o \\
     0 & 0 & 1 \\
    \end{bmatrix}
    \begin{bmatrix} x' \\ y' \\ 1 \end{bmatrix}
\end{equation}
where $\overline{u}=\begin{bmatrix} u & v & 1 \end{bmatrix}^T$ and $\overline{x}'=\begin{bmatrix} x' & y' & 1 \end{bmatrix}^T$ denote homogeneous pixel and homogeneous focal plane normalized coordinates, respectively. The inner matrix
\begin{equation}
    K \triangleq
    \begin{bmatrix}
     f/\mu_x & \gamma & u_o \\
     0 & f/\mu_y & v_o \\
     0 & 0 & 1 \\
    \end{bmatrix}
\end{equation}
defines the camera calibration matrix. 

Matrix $K$ consists of image sensor parameters and intrinsic camera parameters. The image sensor parameters in $K$ are $\mu_x$, $\mu_y$, and $\gamma$. Pixel pitches $\mu_x$ and $\mu_y$ represent the center-to-center distance between adjacent pixels in the image sensor's local $x$ and $y$ directions, respectively \cite{zisserman2004multiple}. Skew angle $\gamma$ expresses the angle between the local $x$ and $y$ directions on the image sensors. Generally the local $x$ and $y$ directions are orthogonal such that $\gamma=0$ for most image sensors, but we include $\gamma$ here for completeness. The intrinsic camera parameters consist of $u_o,v_o,$ and $f$. The principal point $(u_o,v_o)$ provides the location where the camera's boresight intersects the focal plane in pixel coordinates. The effective focal length $f$ defines the focal plane's distance from the origin of the camera coordinate system.




\section{Camera Calibration Algorithm}
Camera manufacturers produce cameras with nominal intrinsic parameters, however these intrinsic parameters are subject to defects and require verification. Camera calibration verifies the intrinsic parameters and replaces the nominal parameter values with their calibrated values. Eq.~\eqref{calib_eq:proportionality} provides an opportunity for estimating $K$ (i.e., calibrating the camera) given a planet image. The spacecraft's state defines $\mathcal{C}$, and image processing supplies $\mathcal{C'}$ from the imaged planet. Eq.~\eqref{calib_eq:proportionality} is a proportionality, but introducing $s$ as the unknown scale coefficient gives
\begin{equation} \label{calib_eq:initial}
    sK^T \mathcal{C'} K = \mathcal{C}
\end{equation}
which converts the proportionality relationship to an equality. Eq.~\eqref{calib_eq:initial} is nonlinear with respect to $K$, and for this reason other works require $N\geq2$ targets/images for estimating $K$ ~\cite{yang2000planar,sun2015accurate,huang2015common,yang2020camera,su2020novel,hu2001camera}.  In the next section, we detail an algorithm that first solves for unknown $s$ and then $K$ for intrinsic camera calibration from the image of a single planet.

\subsection{Matrix Block-Partitioning}
For the camera calibration algorithm, we first partition matrices $K$, $\mathcal{C}$, and $\mathcal{C}'$ into sub-blocks. Block-partitioning of $K$ results in
\begin{equation}
    K \triangleq
    \begin{bmatrix}
    K_{11} & K_{12} \\ 0_{1\times2} & 1\\
    \end{bmatrix}
\end{equation}
where 
\begin{equation}
    \begin{aligned}
        K_{11} \triangleq \begin{bmatrix}
        f/\mu_x & \gamma \\ 0 & f / \mu_y
        \end{bmatrix} \quad \textrm{and} \quad
        K_{12} \triangleq \begin{bmatrix}
        u_o\\ v_o
        \end{bmatrix}.
    \end{aligned}
\end{equation}
Similarly, we'll also block partition $\mathcal{C}$ and $\mathcal{C}'$ as follows
\begin{equation}
    \mathcal{C} \triangleq
    \begin{bmatrix}
    \mathcal{C}_{11} & \mathcal{C}_{12} \\ \mathcal{C}_{12}^T & \mathcal{C}_{22}\\
    \end{bmatrix}, \quad
    \mathcal{C'} \triangleq
    \begin{bmatrix}
    \mathcal{C'}_{11} & \mathcal{C'}_{12} \\ \mathcal{C'}_{12}^T & \mathcal{C'}_{22}\\
    \end{bmatrix}
\end{equation}
where $\mathcal{C}_{11}, \mathcal{C}'_{11} \in \mathbb{R}^{2\times2}$, $\mathcal{C}_{12}, \mathcal{C}'_{12} \in \mathbb{R}^{2\times1}$, and $\mathcal{C}_{22}, \mathcal{C}'_{22} \in \mathbb{R}^{1\times1}$. Substituting the block-partitioned matrices results in 
\begin{equation} \label{calib_eq:blocks}
    \begin{aligned}
            s
    \begin{bmatrix}
    K_{11}^T \mathcal{C'}_{11} K_{11} & K_{11}^T (\mathcal{C'}_{11} K_{12} + \mathcal{C'}_{12} ) \\
    (K_{12}^T \mathcal{C'}_{11} + \mathcal{C'}_{12}^T ) K_{11} & K_{12}^T \mathcal{C'}_{11} K_{12} + 2 K_{12}^T \mathcal{C'}_{12}  + \mathcal{C'}_{22}
    \end{bmatrix}
    =  \\
    \begin{bmatrix}
    \mathcal{C}_{11} & \mathcal{C}_{12} \\
    \mathcal{C}_{12}^T & \mathcal{C}_{22} \\
    \end{bmatrix}.
        \end{aligned}
\end{equation}
We use the system of equations in Eq.~\eqref{calib_eq:blocks} to solve for $s$ directly and then solve for $K_{11}$ and $K_{12}$ separately.

\subsection{Solving for Scale Factor $s$}
When solving for unknown $s$ in systems of equations, Refs.~\cite{christian2021tutorial,modenini2021planet} apply the $trace(\bullet)$ or $det(\bullet)$ operator to isolate $s$. Applying $det(\bullet)$ to the overall system of equations in Eq.~\eqref{calib_eq:initial} results in
\begin{equation} \label{calib_eq:det_whole}
    det(sK^T \mathcal{C'} K) = s^3 det(K)^2 det(\mathcal{C'}) = det(\mathcal{C})
\end{equation}
where $det(\mathcal{C})$ and $det(\mathcal{C}')$ are known but $det(K)$ is not. Interestingly, due to $K$'s upper triangular structure
\begin{equation}
    det(K) = det(K_{11}) det(1) = det(K_{11}).
\end{equation}
such that Eq.~\eqref{calib_eq:det_whole} is also
\begin{equation}
    det(sK^T \mathcal{C'} K) = s^3 det(K_{11})^2 det(\mathcal{C'}) = det(\mathcal{C}).
\end{equation}
Since $K_{11}$ also appears in the sub-block equality
\begin{equation} \label{calilb_eq:K11}
    sK_{11}^T  \mathcal{C'}_{11} K_{11} = \mathcal{C}_{11},
\end{equation}
we also apply the $det(\bullet)$ operator to yield
\begin{equation} \label{calib_eq:K11det}
    s^2 det(K_{11})^2 det(\mathcal{C'}_{11}) = det(\mathcal{C}_{11})
\end{equation}
which is also quadratic in $det(K_{11})$. Since both Eq.~\eqref{calib_eq:det_whole} and Eq.~\eqref{calib_eq:K11det} possess $det(K_{11})^2$ terms, dividing one by the other eliminates the unknown, and re-arranging provides
\begin{equation}
    s = \frac{det(\mathcal{C}) det(\mathcal{C'}_{11})}{det(\mathcal{C'}) det(\mathcal{C}_{11})},
\end{equation}
a direct expression for $s$ in terms of known $\mathcal{C}$, $\mathcal{C}'$, $\mathcal{C}_{11}$, and $\mathcal{C}_{11}'$

\subsection{Solving for $K_{11}$}
With $s$ known, we can now solve for $K_{11}$ and $K_{12}$ to ultimately provide $K$. First we focus on $K_{11}$ in Eq.~\eqref{calilb_eq:K11} which we point out is a symmetric expression quadratic in $K_{11}$. If we can also prove matrices $s\mathcal{C'}_{11}$ and $\mathcal{C}_{11}$ are positive definite, then their Cholesky decompositions
\begin{equation} \label{calib_eq:chol_1}
    \mathcal{C}_{11} = L_\mathcal{C} L_\mathcal{C}^T
\end{equation}
and
\begin{equation} \label{calib_eq:chol_2}
        s \mathcal{C'}_{11} = L_\mathcal{C'} L_\mathcal{C'}^T,
\end{equation}
where $L_{\bullet}$ are lower triangular matrices, reduce Eq.~\eqref{calilb_eq:K11} to
\begin{equation} \label{calib_eq:desired}
    L_\mathcal{C'}^T K_{11} = L_\mathcal{C}^T
\end{equation}
an expression linear in $K_{11}$. This desired linear form provides a direct solution for $K_{11}$.

When assessing the definitions of $s\mathcal{C'}_{11}$ and $\mathcal{C}_{11}$, their $det(\bullet)$ as given in Eq.~\eqref{calib_eq:K11det} is useful. Applying the $sign(\bullet)$ operated defined by
\begin{equation}
    sign(\bullet) 
    \begin{cases}
    \bullet > 0 & sign(\bullet)= +1 \\
    \bullet < 0 & sign(\bullet)=-1\\
    \end{cases}
\end{equation}
to Eq.~\eqref{calib_eq:K11det} results in
\begin{equation} \label{calib_eq:sign_det}
    sign(s^2 det(K_{11})^2 det(\mathcal{C'}_{11})) = sign(det(\mathcal{C}_{11}))
\end{equation}
and provides a means of assessing each term's sign convention. Scale factor $s$ is a real-valued scalar such that $s^2>0$, and therefore $sign(s^2) = +1$. The explicit definition of $det(K_{11})$ 
\begin{equation}
    det(K_{11}) = \frac{f^2}{\mu_x \mu_y} 
\end{equation}
consists of $f$, $\mu_x$, and $\mu_y$ terms that are strictly positive terms. Hence, $det(K_{11}) >0$ by convention, and $sign(det(K_{11})) = +1$. Now substituting $sign(s^2)=+1$ and $sign(det(K_{11})) = +1$ into Eq.~\eqref{calib_eq:sign_det} simplifies to
\begin{equation}
    sign(det(\mathcal{C'}_{11})) = sign(det(\mathcal{C}_{11}))
\end{equation}
in terms of conics $\mathcal{C'}_{11}$ and $\mathcal{C}_{11}$.

It is well known that for ellipses $det(\mathcal{C'}_{11}) > 0$. Given that the ellipsoid's horizon generally projects to an ellipse under perspective projection,
\begin{equation}
    sign(det(\mathcal{C'}_{11})) = sign(det(\mathcal{C}_{11})) = +1
\end{equation}
holds for nearly all viewing configurations of the ellipsoid. Considering $\mathcal{C'}_{11}, \mathcal{C}_{11} \in \mathbb{R}^{2\times2}$ , $\mathcal{C'}_{11}$ and $\mathcal{C}_{11}$ have strictly positive or strictly negative eigenvalues that result in non-negative $det(\bullet)$. In other words, $\mathcal{C'}_{11}$ and $\mathcal{C}_{11}$ are either positive- or negative-definite. To ensure $\mathcal{C'}_{11}$ and $\mathcal{C}_{11}$ are strictly positive-definite, we modify $\mathcal{C'}$ and $\mathcal{C}$ to 
\begin{equation}
\mathcal{C'} = \alpha \mathcal{C'} \\
\end{equation}
and 
\begin{equation}
    \mathcal{C} = \beta \mathcal{C} \\
\end{equation}
where $\alpha=sign(trace(\mathcal{C'}_{11}))$ and $\beta=sign(trace(\mathcal{C}_{11}))$. As $2\times2$ matrices, non-negative $trace(\bullet)$ implies both eigenvalues are positive (i.e., positive definite) and vice-versa for non-positive $trace(\bullet)$. With this modification, $s\mathcal{C}_{11}$ and $\mathcal{C'}_{11}$ are symmetric positive definite which enables their Cholesky decompositions $L_\mathcal{C}$ and $L_\mathcal{C'}$ given by Eq.~\eqref{calib_eq:chol_1} and Eq.~\eqref{calib_eq:chol_2}, respectively.

Substituting $L_\mathcal{C'}$ and $L_\mathcal{C}$ into Eq.~\eqref{calilb_eq:K11} produces
\begin{equation}
    K_{11}^T L_\mathcal{C'} L_\mathcal{C'}^T K_{11} = L_\mathcal{C} L_\mathcal{C}^T
\end{equation}
from which we obtain the desired form in Eq.~\eqref{calib_eq:desired}. Re-arranging and solving for $K_{11}$ provides
\begin{equation} \label{calib_eq:K11_cholesky}
    K_{11} = L_\mathcal{C'}^{-T} L_\mathcal{C}^T
\end{equation}
in exact terms. 

\subsection{Solving for $K_{12}$}

With $K_{11}$ known, we observe the following sub-block equality
\begin{equation} \label{calib_eq:K12}
    sK_{11}^T (\mathcal{C'}_{11} K_{12} + \mathcal{C'}_{12}) = \mathcal{C}_{12}
\end{equation}
is linear in $K_{12}$ and contains known terms $K_{11},s,\mathcal{C'}_{11},\mathcal{C'}_{12},$ and $\mathcal{C}_{12}$. Re-arranging Eq.~\eqref{calib_eq:K12} to
\begin{equation} \label{calib_eq:principal}
    K_{12} = \mathcal{C'}_{11}^{-1} \left( (sK_{11}^T)^{-1} \mathcal{C}_{12} - \mathcal{C'}_{12} \right)
\end{equation}
provides a direct solution for $K_{12}$. Substituting Eq.~\eqref{calib_eq:K11_cholesky} and Eq.~\eqref{calib_eq:chol_2} simplifies Eq.~\eqref{calib_eq:principal} to
\begin{equation} \label{calib_eq:principal_simple}
    K_{12} = (L_{\mathcal{C}} L_{\mathcal{C'}}^{T} )^{-1} \mathcal{C}_{12} - \mathcal{C'}_{11}^{-1}\mathcal{C'}_{12}
\end{equation}
consisting solely of terms involving $s$, $\mathcal{C}$, and $\mathcal{C'}$. For ease of notation, we define 
\begin{equation}
    \mathcal{J} \triangleq (L_{\mathcal{C}} L_{\mathcal{C'}}^{T} )^{-1} \mathcal{C}_{12} - \mathcal{C'}_{11}^{-1}\mathcal{C'}_{12}
\end{equation}
as the right-hand side of Eq.~\eqref{calib_eq:principal_simple}.  With $K_{11}$ and $K_{12}$ known, our algorithm estimates the camera calibration matrix $K$ from a single imaged ellipsoid.

\subsection{Camera Calibration Algorithm Summary}

\begin{algorithm} 
\SetKw{Next}{next}
\SetKw{Procedure}{procedure}
\SetKwFunction{Function}{function}
\KwIn{ $\mathcal{C'}$, $\mathcal{C}$ }
\KwOut{ $K$ }
\DontPrintSemicolon
\SetAlgoLined
compute $\alpha$ = $sign(trace(\mathcal{C'}_{11}))$ \;
compute $\beta$ = $sign(trace(\mathcal{C}_{11}))$ \;
update $\mathcal{C'} = \alpha \mathcal{C'}$ \;
update $\mathcal{C} = \beta \mathcal{C}$ \;
compute $s = det(\mathcal{C})det(\mathcal{C'}_{11})/ \left(det(\mathcal{C'})det(\mathcal{C}_{11}) \right)$ \;
compute $L_\mathcal{C'} = chol(s \mathcal{C'}_{11})$ \;
compute $L_\mathcal{C} = chol(\mathcal{C}_{11})$ \;
compute $K_{11} = L_{\mathcal{C'}}^{-T} L_{\mathcal{C}}^{T}$ \;
compute $K_{12}$ from $K_{12} = \mathcal{J}$ \;
build $K$ from $K_{11}$ and $K_{12}$ \;
\caption{Pseudocode for Camera Calibration Matrix Estimate}
\label{algo:calibration_original}
\end{algorithm}

The camera calibration algorithm from a single imaged ellipsoid is quite simple. Algorithm~\ref{algo:calibration_original} details the entire algorithm in 10 lines of pseudo-code where $\mathcal{C}$ and $\mathcal{C'}$ are inputs. For each line of pseudo-code, we analytically compute the required floating point (FLOP) count and report it in Table~\ref{calib_tab:flop}. We refer to Refs.~\cite{golub1996matrix,trefethen1997numerical} for the appropriate FLOP count approximations for the algorithm lines involving Cholesky factorization and triangular matrix inversions. In summary, the camera calibration algorithm provides an estimate for $K$ from a single imaged ellipsoid in $\sim 135$ FLOPs which for context is slightly more than the FLOPs required to invert a $5\times5$ matrix (i.e., $\sim 125$ FLOPS) \cite{golub1996matrix,trefethen1997numerical}.

\begin{table}[htbp!]
    \centering
        \caption{Camera Calibration Algorithm FLOP Count}
        
    \begin{tabular}{cc}
        \hline \hline Line & FLOPs \\ \hline
        1 & 3 \\
        2 & 3 \\
        3 & 9 \\
        4 & 9 \\
        5 & 41 \\
        6 & 7 \\
        7 & 3 \\
        8 & 18 \\ 
        9 & 42  \\ \hline
        Total & 135 \\ \hline \hline
    \end{tabular}
    \label{calib_tab:flop}
\end{table}

\section{Algorithm Extensions}
The proposed camera calibration algorithm estimates $K$ from a single imaged ellipsoid which applies to planetary images. A camera's focal length $f$ is an important intrinsic parameter that is embedded within $K$ but requires knowledge of the camera's image sensor. In this section we provide $f$ estimation from $K$. Additionally, our camera calibration algorithm extends to multiple images for estimating $K$ and the intrinsic parameters in a least-squares or batch-filter approach. In this section we also extend the algorithm for batch-filter estimates from multiple images.

\subsection{Focal Length Estimation}
Within $K$, the camera's focal length $f$ appears in the $f/\mu_x$ and $f/\mu_y$ terms. Introducing standard basis vectors $e_1\triangleq \begin{bmatrix} 1 & 0\end{bmatrix}^T$ and $e_2\triangleq \begin{bmatrix} 0 & 1\end{bmatrix}^T$, we extract the $f$ terms from estimated $K_{11}$ through
\begin{equation}
    d_x \triangleq e_1^T K_{11} e_1 = f/\mu_x
\end{equation}
and
\begin{equation}
    d_y \triangleq e_2^T K_{11} e_2  = f/\mu_y
\end{equation}
where we introduce $d_x$ and $d_y$ for convenience.
Through directly estimating $K$, it is impossible to estimate $f$ without first knowing the image sensor's pixel pitches $\mu_x$ and $\mu_y$ \cite{christian2021tutorial}. Therefore, using the image sensor's $\mu_x$ and $\mu_y$, the following linear system
\begin{equation} \label{calib_eq:focal}
    \begin{bmatrix}
    1 \\ 1
    \end{bmatrix}
    f = 
    \begin{bmatrix}
    \mu_x d_x \\ \mu_y d_y
    \end{bmatrix}
\end{equation}
estimates $f$ in a least-squares sense from estimated $K$. Even from a single imaged planet in a single image, Eq.~\eqref{calib_eq:focal} provides an over-determined system of equations for solving $f$.

\subsection{Extension to Multi-Image Calibration}
Though we prove camera calibration is possible from a single image, extending the calibration to multiple images enables more accurate and precise estimates for the camera's intrinsic parameters. Eq.~\eqref{calib_eq:focal} is an over-determined system from a single image. We introduce $d_{x,i}$ and $d_{y,i}$ to denote the $d_x$ and $d_y$ values from the $i^{th}$ image so that
\begin{equation} \label{calib_eq:focal_multiple}
    \begin{bmatrix}
    1 \\
    1 \\
    \vdots \\
    1 \\
    1 \\    
    \end{bmatrix}
    f = 
    \begin{bmatrix}
    \mu_x d_{x,1} \\ 
    \mu_y d_{y,1} \\
    \vdots \\
    \mu_x d_{x,N} \\ 
    \mu_y d_{y,N} \\
    \end{bmatrix}
\end{equation}
augments Eq.~\eqref{calib_eq:focal} to $f$ estimation from multiple images. Since each image provides $2$ entries to Eq.~\eqref{calib_eq:focal_multiple}, the uncertainty in $f$ estimates scales with $\sim 1 / \sqrt{2N}$ where $N$ is the total number of images.

When estimating $K$, sub-block $K_{12}$ provides an estimate for the principal point's coordinates $(u_o,v_o)$ directly. Augmenting the principal point estimate using multiple images leads to 

\begin{equation} \label{calib_eq:principal_multiple}
    \begin{bmatrix}
    I_{2\times2} \\
    \vdots \\
    I_{2\times2}\\
    \end{bmatrix}
    \begin{bmatrix}
    u_o \\ v_o
    \end{bmatrix} =
    \begin{bmatrix}
    \mathcal{J}_1\\
    \vdots \\
    \mathcal{J}_N\\
    \end{bmatrix}
\end{equation}
a least squares problem where $\mathcal{J}_i$ is the $\mathcal{J}$ matrix of the $i^{th}$ image. Matrix $I_{2\times2}$ has dimensions $2\times2$ such that uncertainty in $(u_o,v_o)$ scales with $\sim 1 / \sqrt{N}$.



\section{Numerical Simulations}
\begin{figure*}[htb!]
\centering
\subfloat[Focal Length Estimate Uncertainty with Increasing Images]{ \includegraphics[width=0.8\textwidth]{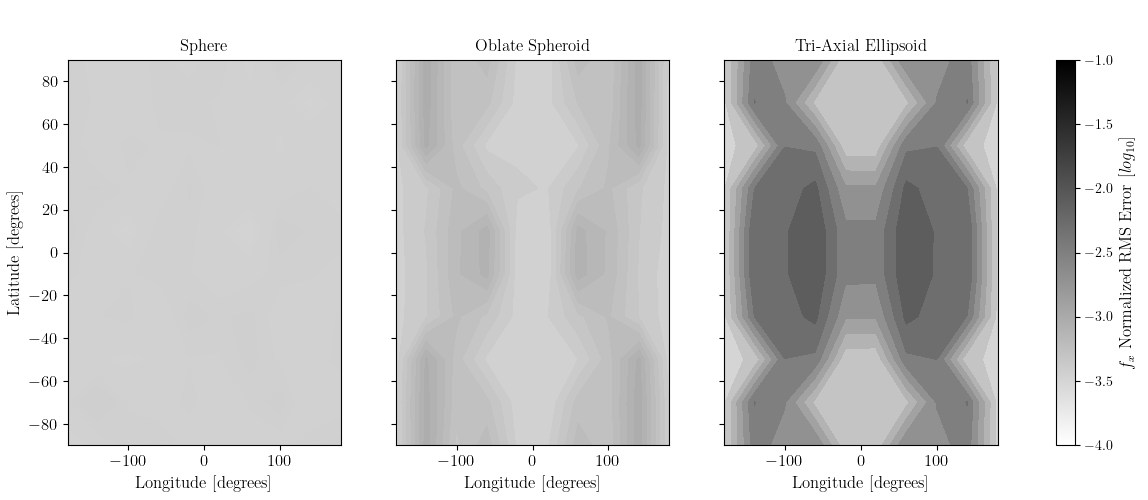}
    \label{fig:NRMS_focal}}
\hfill
\subfloat[Principal Point Estimate Uncertainty with Increasing Images]{\includegraphics[width=0.8\textwidth]{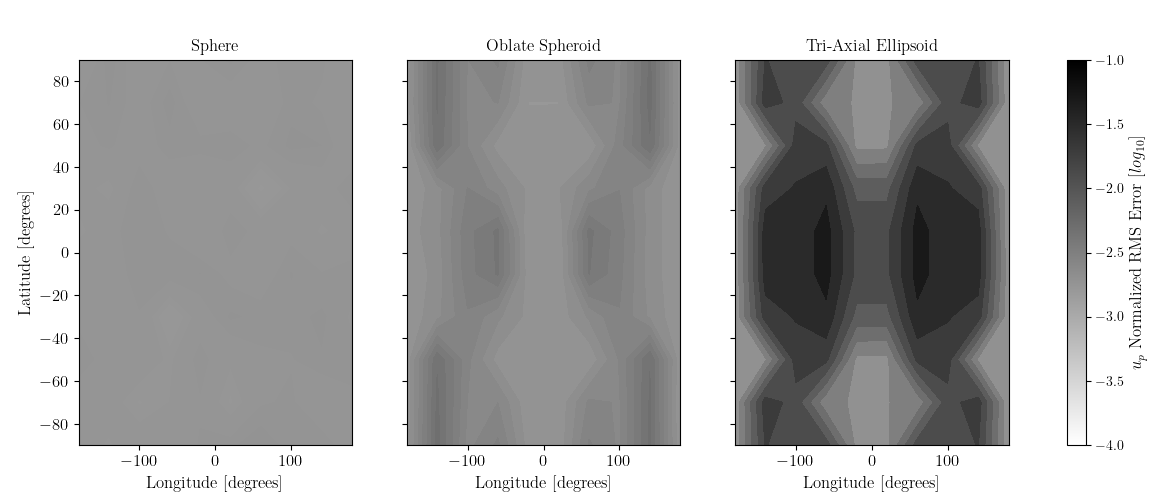}%
    \label{fig:NRMS_center}}
\caption{Performance of the Proposed Camera Calibration Algorithm for Varying Simulated Ellipsoids and Varying Poses (a) Uncertainty in $f_x$ as Reported by NRMS Metric (b) Uncertainty in $u_o$ as Reported by NRMS Metric}
\end{figure*}

We simulate planetary bodies of varying ellipsoid shapes and pointing geometries to assess our method's camera calibration performance. Table~\ref{tab:ellipsoid shape} details the semi-axes values used to model each ellipsoid in terms of the planet's polar radius $R_p$. 
\begin{table}[htbp!]
    \centering
    \caption{Simulated Ellipsoid Semi-axes}
    \begin{tabular}{ccccc}
    \hline \hline
        Ellipsoid-Type & $a \ (R_p)$ & $b  \ (R_p)$ & $c  \ (R_p)$ \\ \hline
        Sphere & 1.0 & 1.0 & 1.0 \\
        Oblate Spheroid & 1.0 & 1.5 & 1.5 \\
        Triaxial Ellipsoid & 1.0 & 2.0 & 3.0 \\
        \hline \hline
    \end{tabular}
    \label{tab:ellipsoid shape}
\end{table}
We examine the method's sensitivity to ellipse-fit error by perturbing the semi-major/minor axes and center coordinates of the imaged ellipse $\mathcal{C'}$ with Gaussian noise $\sim \mathcal{N}(0,\sigma^2)$. Here the proportionality relationship 
\begin{equation}
    \mathcal{C'} \propto K^{-T} \mathcal{C'} K^{-1}
\end{equation}
gives the imaged ellipse $\mathcal{C'}$ in pixel coordinates~\cite{christian2021tutorial}, similar to what an observer computes. The Gaussian noise perturbation of $1 \sigma = 1$ pixels models effects of edge localization error typical of off-the-shelf edge detection algorithms~\cite{christian2017accurate}. The $1\sigma =1$ pixel perturbation serves as a large, conservative perturbation given that semi-major/minor axes and center coordinates are usually known to sub-pixel precision for a fitted ellipse. After perturbation, we compare the estimated intrinsic parameters with the ground truth and obtain the residual. 

The simulation performs a pose-varying Monte Carlo (MC) simulation for a nadir-pointing (i.e., pointing at the planet center) spacecraft. We sample all possible viewing latitudes (i.e., $-90^\circ$ to $90 ^\circ$) and longitudes (i.e., $-180^\circ$ to $180^\circ$) at a distance of $10 R_p$ with a $10\times 10$ grid and obtain the root-mean-square (RMS) of 1000 MC runs per gridpoint. To generalize the findings, we divide the RMS value by the ground-truth value to obtain the normalized root-mean-square (NRMS). Figure~\ref{fig:NRMS_focal} and Fig.~\ref{fig:NRMS_center} report the NRMS of $f$ and $(u_o,v_o)$, respectively, for
varying ellipsoid shapes and viewing poses of a nadir-pointing spacecraft. 

From Fig.~\ref{fig:NRMS_focal} and Fig.~\ref{fig:NRMS_center} the proposed method estimates $f$ with greater precision than $(u_o,v_o)$ as witnessed by the lighter contours. Since $K_{12}$ is computed from $K_{11}$, the estimation error of $K_{11}$ propagates to the $(u_o,v_o)$ estimates. Additionally, Fig.~\ref{fig:NRMS_focal} and Fig.~\ref{fig:NRMS_center} also illustrate the effect the shape of the apparent horizon has on estimated $f$ and $(u_o,v_o)$. Under the nadir-pointing assumption, the apparent horizon of a sphere is a circle for all poses, and thus all poses yield similar camera calibration performance. However when semi-axes $b \neq a$ and $c \neq a$ as in the oblate spheroid and triaxial ellipsoid cases, the observer's pose influences the shape of the apparent horizon . The apparent horizon is no longer a circle but an ellipse with arbitrary eccentricity. As witnessed by the oblate spheroid and triaxial ellipsoid cases in Fig.~\ref{fig:NRMS_focal} and Fig.~\ref{fig:NRMS_center}, the pose-dependent apparent horizon results in varying degrees of camera calibration performance. The NRMS contours of $f$ and $(u_o,v_o)$ provide insight on what an observer's pose needs to be for a desired level of camera calibration performance from an imaged ellipsoid.


\section{Application to Planetary Images}
\subsection{Image Dataset}
To verify our simulated findings, we subject the camera calibration algorithm to planetary images. As Ref.~\cite{christian2017accurate} points out, horizon-based navigation works best with ellipsoidal bodies without an atmosphere given that the ellipsoid does not model the atmosphere. For this reason, we select the \textit{Cassini} Imaging Science Subsystem (ISS) dataset (available through NASA Planetary Data System) that contains numerous images of Saturn's atmosphere-less, ellipsoidal moons as imaged by the spacecraft \textit{Cassini}. \textit{Cassini} was a spacecraft that explored and studied the Saturn system.  The \textit{Cassini} ISS consists of a Narrow Angle Camera (NAC) and Wide Angle Camera (WAC) with $0.35^\circ \times 0.35^\circ $ and $3.5^\circ \times 3.5^\circ $ field of view, respectively \cite{porco2004cassini}. Out of the two cameras, we subject our algorithm to NAC images due to its higher angular resolution. Within the ISS NAC dataset, the Saturnian ellipsoidal moons listed in Table~\ref{tab:ellipsoidal_moons} serve as the imaging targets for camera calibration. Table~\ref{tab:ellipsoidal_moons} also lists the semi-major axes for each moon's best-fit ellipsoid. In total, our ISS NAC dataset consists of $50$ planetary images.

\begin{table}[htbp!]
    \centering
    \caption{Saturnian Moons' Ellipsoid Semi-axes~\cite{dougherty2009saturn,acton1996ancillary,acton2018look}}
    \begin{tabular}{ccccc}
    \hline \hline
        Moon & $a \ (km)$ & $b  \ (km)$ & $c  \ (km)$ \\ \hline
        Mimas & 415.6 & 393.4 & 381.2 \\
        Tethys & 1076.8 & 1057.4 & 1052.6 \\
        Enceladus & 513.2 & 502.8 & 496.6 \\
        Iapetus & 1492.0 & 1492.0 & 1424.0 \\
        Rhea & 1532.4 & 1525.6 & 1524.4 \\
        Dione & 1128.8 & 1122.6 & 1119.2 \\
        \hline \hline
    \end{tabular}
    \label{tab:ellipsoidal_moons}
\end{table}

\subsection{Image Processing}
\begin{figure*}[tbp!]
    \centering
    \includegraphics[width=\textwidth]{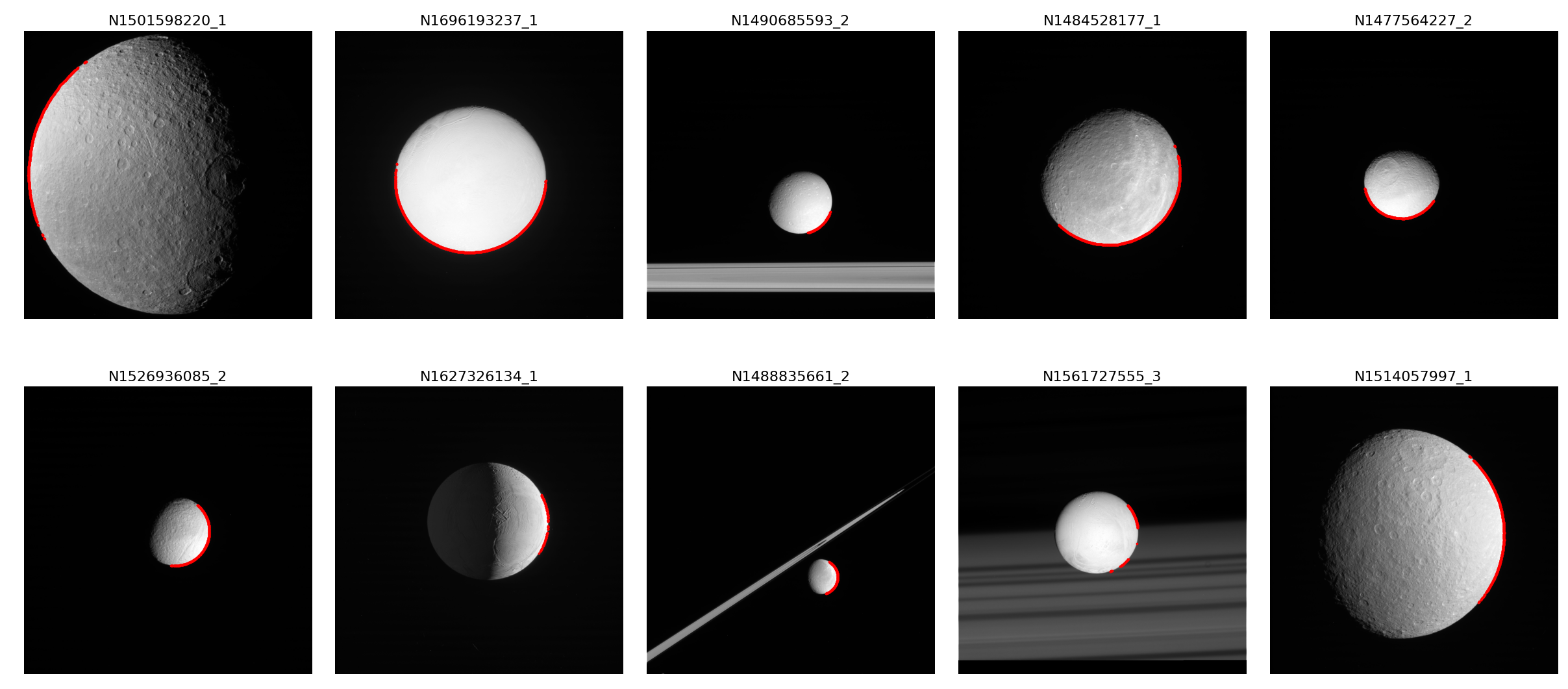}
    \caption{A Subset of the Processed Images in the \textit{Cassini} ISS NAC Dataset with their respective NASA Planetary Data System ID above the Image. The rings of Saturn appear in the background for some.}
    \label{fig:cassinii_nac_all}
\end{figure*}
For camera calibration, our algorithm requires extracting the imaged horizon conic from the imaged ellipsoid. We obtain the imaged conic by first extracting the apparent horizon points in the image and then fitting the extracted points to a conic section. For horizon point extraction, we employ the partial area effect algorithm in Ref.~\cite{trujillo2013accurate} for subpixel edge estimates. Typically, limb scanning precedes subpixel edge estimation and provides a pixel-level guess of the planetary body's limb prior to refinement via subpixel edge estimation. However in our application, subpixel estimation without limb scanning results in accurate horizon extraction for the vast majority of images. Figure~\ref{fig:cassinii_nac_all} provides an example of the extracted horizon conic at the subpixel level for 10 images from the ISS NAC dataset used in this work.

With accurate subpixel estimates of the extracted horizon, we apply Ref.~\cite{kanatani2011hyper}'s Semi-Hyper Least-Squares algorithm for an unbiased, direct conic-fit. The conic-fit's coefficients assemble to form the imaged conic $\mathcal{C'}$.

\subsection{Results}
From each of the $50$ images in the \textit{Cassini} ISS NAC dataset, our algorithm estimates $K$ for \textit{Cassini}'s NAC. We compare our estimated focal length $f$ and principal point $(u_o,v_o)$ intrinsic parameters to their documented calibrated values to assess our algorithm's performance. Camera calibration for the \textit{Cassini} ISS NAC consisted of a ground calibration segment and an in-orbit calibration segment. Ground-based calibration provides nominal values for the intrinsic parameters, and the in-orbit calibration segment corrects these nominal values with higher accuracy/precision  through imaging star clusters \cite{porco2004cassini}.

\subsubsection{Focal Length}
\begin{figure}[htbp!]
    \centering
    \includegraphics[width=0.5\textwidth]{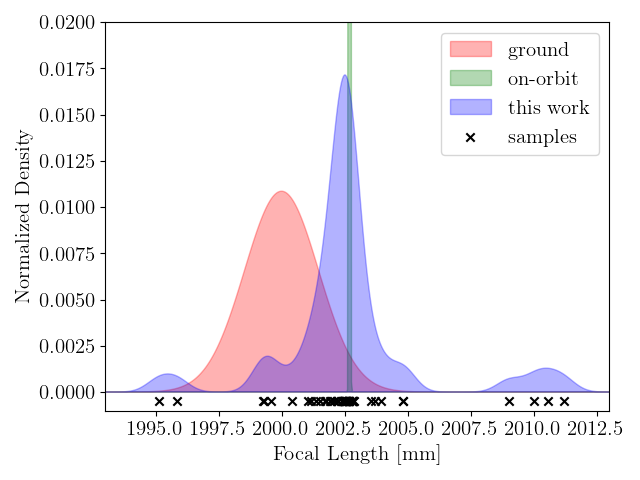}
    \caption{Distributions of $f$ Estimates}
    \label{fig:focal_distribution}
\end{figure}

From each image's $K$ estimate, appying Eq.~\eqref{calib_eq:focal} provides $f$. The $f$ estimates of each image provides a sample population from which we report its central tendency and statistical dispersion in Table~\ref{tab:method_focals}. For context, Table~\ref{tab:cassini_nac_focals} presents the calibrated $f$ values for the \textit{Cassini} ISS NAC where the in-orbit $f$ serves as the ground-truth value for comparison. Since we did not employ limb scanning, our image processing also extracts additional edges that do not belong to planet's horizon and biases the fitted conic. The biased conic-fit then leads to biased $K$ estimates and produces a handful of outliers. The robust statistical measures such as $median(\bullet)$ and median absolute deviation $MAD(\bullet)$ are robust to outliers and provide insight on what to expect had we included limb scanning in the image processing. We illustrate the different distributions in $f$ estimates in Fig.~\ref{fig:focal_distribution}.

\begin{table}[htbp!]
    \centering
    \caption{\textit{Cassini} ISS NAC Calibrated Focal Length Values \cite{porco2004cassini}}
    \begin{tabular}{ccc}
    \hline \hline
        Metric & Ground-Calibrated & On-Orbit \\ \hline
    Mean $(mm)$  & 2000.0 & 2002.7  \\
    $3\sigma \ (mm)$ & 4.0 & 0.07  \\
        \hline \hline
    \end{tabular}
    \label{tab:cassini_nac_focals}
\end{table}

\begin{table}[htbp!]
    \centering
    \caption{Proposed Algorithm's Estimated Focal Length Statistics}
    \begin{tabular}{cc}
    \hline \hline
        Metric & Units $(mm)$ \\ \hline
        Mean Value & 2002.7\\
        Median Value & 2002.5 \\
    Mean Error & 0.04 \\
    Median Error & 0.18 \\
    $3\sigma$ & 8.4  \\
    $3MAD$ & 0.9  \\
        \hline \hline
    \end{tabular}
    \label{tab:method_focals}
\end{table}


Fig.~\ref{fig:focal_distribution} plots the distributions of \textit{Cassini}'s $f$ estimates according to their respective normal distribution parameters reported in Ref.~\cite{porco2004cassini,knowles2016cassini}. Fig.~\ref{fig:focal_distribution} plots the $f$ estimates for each image in the dataset using our algorithm and then applies kernel density estimation with a Gaussian kernel to visualize the sampled distribution. Treating the on-orbit calibrated $f$ value as the truth, our method provides a more accurate and precise estimate of $f$ when compared to the ground calibrated $f$ value as seen by the spread and peak of our method's distribution. The $mean(\bullet)$ and $median(\bullet)$ metrics confirm our method's central tendency agrees with the true $f$, and the $MAD(\bullet)$ confirms the lower statistical spread compared to the ground-calibrated value of $f$. Our method is essentially as accurate as the on-orbit calibration method for $f$, but not as precise as evidenced by the larger statistical spread.

\subsubsection{Principal Point}
\begin{figure}[htbp!]
    \centering
    \includegraphics[width=0.5\textwidth]{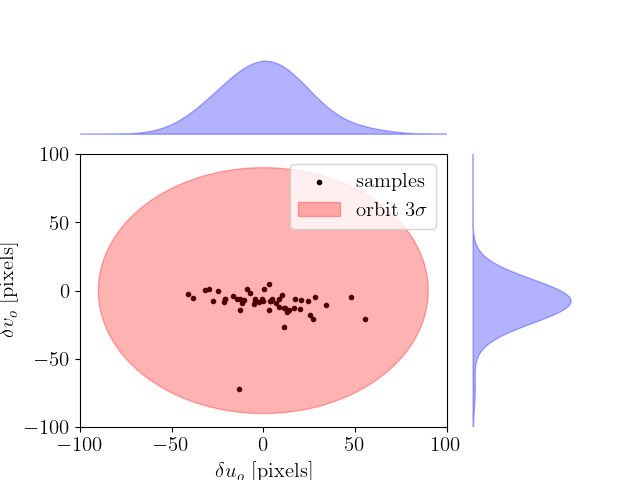}
    \caption{Proposed Algorithm's Principal Point Residuals}
    \label{fig:principal_residuals}
\end{figure}

We continue the comparison with estimates for \textit{Cassini} ISS NAC's principal point $(u_o,v_o)$. Once again, Eq.~\eqref{calib_eq:principal} provides our algorithm's estimate for $(u_o,v_o)$. Table~\ref{tab:cassini_principal} and Table~\ref{tab:method_principal} report the values and statistics of \textit{Cassini} ISS NAC's in-orbit calibrated values of $(u_o,v_o)$ \cite{knowles2016cassini,porco2004cassini} and those from our method, respectively. Holding the in-orbit calibrated values for $(u_o,v_o)$ as truth, Fig.~\ref{fig:principal_residuals} plots the residuals of our method's estimates compared to the truth as well as the $3\sigma$ ellipse from in-orbit calibration. The sides of Fig.~\ref{fig:principal_residuals} illustrate the distribution sampled from the residuals through kernel density estimation with a Gaussian kernel.
\begin{table}[htbp!]
    \centering
    \caption{\textit{Cassini} ISS NAC Calibrated Principal Point Values \cite{porco2004cassini}}
    \begin{tabular}{ccc}
    \hline \hline
        Metric & $u_o$ $(pixels)$ & $v_o$ $(pixels)$ \\ \hline
        Mean & 560 & 500\\
        $1 \sigma$ & 30 & 30 \\
        \hline \hline
    \end{tabular}
    \label{tab:cassini_principal}
\end{table}
\begin{table}[htbp!]
    \centering
    \caption{Proposed Algorithm's Estimated Principal Point Values}
    \begin{tabular}{ccc}
    \hline \hline
        Metric & $u_o$ $(pixels)$ & $v_o$ $(pixels)$ \\ \hline
        Mean Value & 558. 97 & 509.36 \\
        Median Value & 558. 17 & 507.20 \\
        Mean Error & 1.03 & -9.36 \\
        Median Error & 1.83 & -7.21 \\
        $1\sigma$ & 20.48 & 10.80 \\
        $1 MAD$ & 14.21 & 3.08 \\
        \hline \hline
    \end{tabular}
    \label{tab:method_principal}
\end{table}
As evidenced in Fig.~\ref{fig:principal_residuals}, our method's residuals are centered about the in-orbit calibration's $3\sigma$ with all residual samples within of it. Our method provides $(u_o,v_o)$ estimates with higher certainty than the in-orbit calibration method by at least $\sim 10$ pixels as demonstrated by the $1\sigma$ value in Table~\ref{tab:method_principal}.

\begin{figure*}[!t]
\centering
\subfloat[Focal Length Estimate Uncertainty with Increasing Images]{ \includegraphics[width=0.475\textwidth]{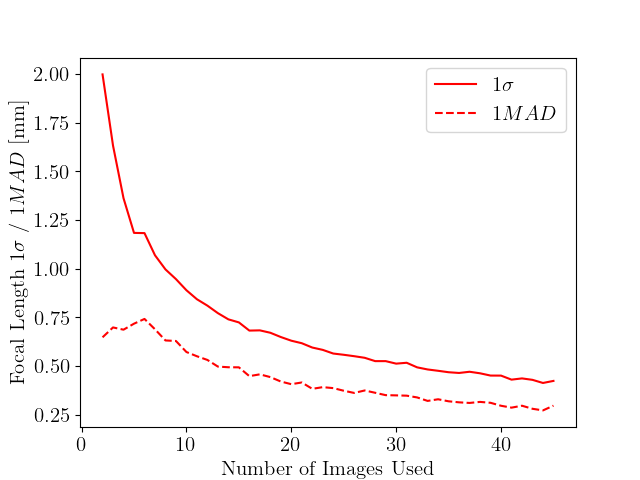}
    \label{fig:multiple_focal}}
\hfill
\subfloat[Principal Point Estimate Uncertainty with Increasing Images]{\includegraphics[width=0.475\textwidth]{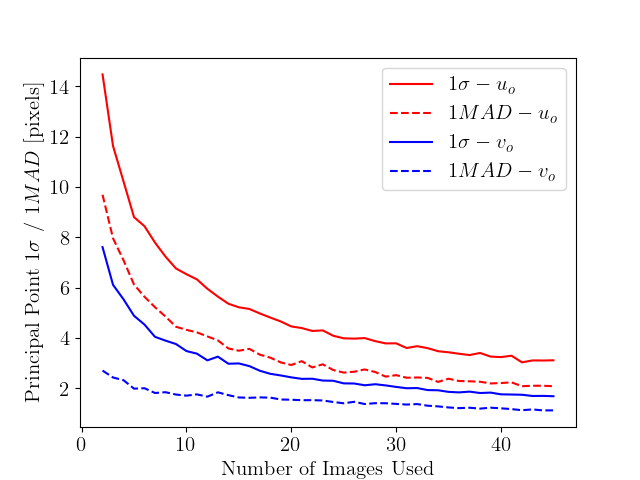}%
\label{fig:multiple_optical}}
\caption{Estimate Uncertainty for the Mulitple Image Extension of the Proposed Calibration Algorithm: (a) Uncertainty in $f$; (b) Uncertainty in $(u_o,v_o)$}
\end{figure*}

\subsection{Extension to Multiple Images}
Figure~\ref{fig:focal_distribution} and Fig.~\ref{fig:principal_residuals} compare our camera calibration algorithm with the star cluster-based calibration. However, this is not a fair comparison between the two due to the amount of images each method requires. For instance, our method requires a single image for a $K$ estimate, while the star-cluster calibration employs multiple images. Ref.~\cite{westCassiniISS} details the calibration test procedures for in-orbit \textit{Cassini} ISS NAC calibration and requires $450$ total images of the star cluster targets for NAC calibration. For proper comparison we now apply the multiple image extension of our algorithm given by Eq.~\eqref{calib_eq:focal_multiple} and Eq.~\eqref{calib_eq:principal_multiple} for multiple images.

\begin{figure}[b!]
    \centering
    \includegraphics[width=0.5\textwidth]{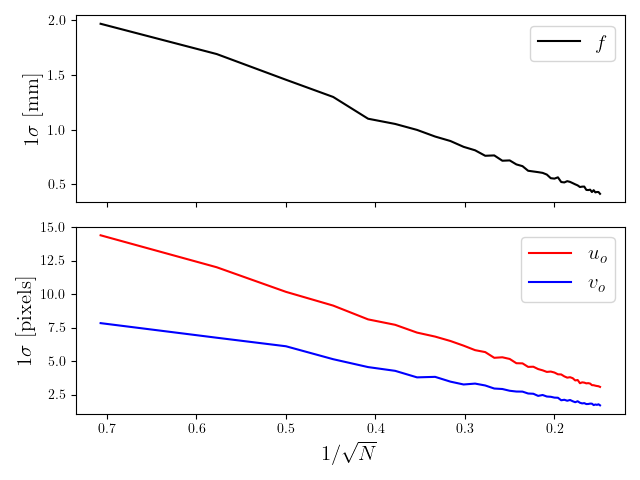}
    \caption{Principal Point $1\sigma$ Uncertainty $f$ and $(u_o,v_o)$ with Respect to $1/ \sqrt{N}$}
    \label{fig:sqrt_N_all}
\end{figure}

Our \textit{Cassini} ISS NAC dataset contains 50 processed images, but to simulate multi-image calibration we'll sample multiple combinations of $q$ images from the total $n=50$ images. For each $q$, 
\begin{equation}
    {}_nC_q = 
    \begin{pmatrix}
    n \\ q
    \end{pmatrix}
    =
    \frac{n!}{q!(n-q)!}
\end{equation}
provides the total unique image combinations possible in the existing \textit{Cassini} ISS NAC dataset. We sample $2000$ combinations for each $q$ and report the resultant $1\sigma$ and $1MAD$ for $f$ and $(u_o,v_o)$ estimates in Fig.~\ref{fig:multiple_focal} and Fig.~\ref{fig:multiple_optical}, respectively. As expected, the uncertainty in both estimates denoted by $1\sigma$ and $1MAD$ decrease with increasing images. As a robust statistic, $1 MAD$ serves as the lower bound of the statistical dispersion of the estimates, and $1\sigma$ serves as the upper bound. From our simulations, at $q=45$ images we expect $f$ uncertainty to lie between $0.30 - 0.43$ mm and $(u_o,v_o)$ uncertainty bounded by $1.1-3.1$ pixels. Recalling the ground truths in Table~\ref{tab:cassini_nac_focals} and Table~\ref{tab:cassini_principal}, multi-image calibration improves $f$ uncertainty such that it approaches the star cluster-calibration uncertainty but with far fewer images (about an order of magnitude fewer). Additionally, at $q=45$ precision in our $(u_o,v_o)$  surpasses the star cluster-calibrated precision by about one order of magnitude.  Both focal length and principal point estimate uncertainties scale  $\sim 1 / \sqrt{N}$ with increasing $N$. 
Fig.~\ref{fig:sqrt_N_all} illustrates the trends linear in $ 1 / \sqrt{N}$ for $1\sigma$ uncertainties in $f$ and $(u_o,v_o)$.


\section{Discussion}
Our camera calibration algorithm is the first calibration algorithm that estimates $K$ from a single imaged ellipsoid. Since ellipsoids are good shape models for planets and moons, our algorithm enables calibration from the nearest ellipsoidal planetary body from a single image. Using \textit{Cassini} ISS NAC as a case study, our algorithm estimates $f$ to higher accuracy and precision when compared to the ground calibration. Though as equally accurate, our algorithm is not nearly as precise at the in-orbit calibrated $f$ values using star clusters. Conversely for principal point $(u_o,v_o)$ estimation, our algorithm is more precise than the star-cluster calibration method and just as accurate for $u_o$ estimates. A slight $\sim 7$ pixel bias exists for the $v_o$ estimate. The multi-image extension to our algorithm improves the precision of both $f$ and $(u_o,v_o)$ estimates. At $q=45$ images, uncertainty in our algorithm's outputs approach and surpass that of the star-calibration method for $f$ and $(u_o,v_o)$ estimates, respectively, despite using $\sim 1$ order of magnitude fewer images. With increased $q$, we expect our algorithm's precision to surpass that of the star cluster-calibration method for both $f$ and $(u_o,v_o)$ estimates.

Though formulated with planets in mind, our algorithm applies to existing camera calibration setups documented in Refs.~\cite{yang2000planar,sun2015accurate,huang2015common,hu2001camera} where a single image captures numerous ellipsoids. The multi-image extensions in Eq.~\eqref{calib_eq:focal_multiple} and Eq.~\eqref{calib_eq:principal_multiple} also apply to multiple ellipsoids captured within the same image. Each imaged ellipsoid provides estimates for $f$ and $(u_o,v_o)$ from which Eq.~\eqref{calib_eq:focal_multiple} and Eq.~\eqref{calib_eq:principal_multiple} provide optimal estimates in a least-squares sense.

\section{Conclusion}
This work provides a novel camera calibration method requiring only one imaged ellipsoid and enables camera calibration using the nearest ellipsoidal planetary bodies. For spacecraft in Earth orbit, the spacecraft need not look further than the moon for calibrating its onboard camera. Our algorithm estimates the camera calibration matrix $K$ from a single image but extends to multiple images for higher precision estimates. The algorithm also applies to ground-based calibration where only a single ellipsoidal imaging target is required.


%

\appendices
\section{Images used in Cassini ISS NAC Dataset}
The NASA Planetary Data System is an online repository that archives the data collected of NASA planetary missions such as images from \textit{Cassini}'s NAC. The \textit{Cassini} ISS NAC dataset used in this work consists of the 50 images listed in Table~\ref{tab:cassini_iss_nac_images}. Each of these images are $1024$ pixels $\times 1024$ pixels in size and corrected for radial distortions.

\begin{table}[htbp!]
    \centering
        \caption{Images used in \textit{Cassini} ISS NAC Dataset}
    \begin{tabular}{cc}
    \hline \hline Image Name & Image Name \\ \hline
\texttt{N1477564227\_2\_full.png} &
\texttt{N1484509816\_1\_full.png} \\
\texttt{N1484528177\_1\_full.png} &
\texttt{N1487418843\_1\_full.png} \\
\texttt{N1488822445\_1\_full.png} &
\texttt{N1488823675\_1\_full.png} \\
\texttt{N1488835661\_2\_full.png} &
\texttt{N1488913320\_4\_full.png} \\
\texttt{N1489254532\_1\_full.png} & 
\texttt{N1490685593\_2\_full.png} \\ 
\texttt{N1495319194\_1\_full.png} & 
\texttt{N1496418930\_2\_full.png} \\
\texttt{N1498348607\_1\_full.png} &
\texttt{N1498573664\_2\_full.png} \\ 
\texttt{N1500045859\_2\_full.png} &
\texttt{N1501598220\_1\_full.png} \\
\texttt{N1501611723\_1\_full.png} &
\texttt{N1501630084\_1\_full.png} \\
\texttt{N1505552943\_1\_full.png} &
\texttt{N1506051828\_1\_full.png} \\
\texttt{N1506110094\_1\_full.png} &
\texttt{N1507577380\_2\_full.png} \\
\texttt{N1507717036\_1\_full.png} & 
\texttt{N1508930209\_2\_full.png} \\
\texttt{N1511700120\_1\_full.png} &
\texttt{N1514057997\_1\_full.png} \\
\texttt{N1514074610\_1\_full.png} &
\texttt{N1514129966\_1\_full.png} \\
\texttt{N1519488890\_1\_full.png} &
\texttt{N1526936085\_2\_full.png} \\ 
\texttt{N1558924482\_3\_full.png} & 
\texttt{N1561727555\_3\_full.png} \\ 
\texttt{N1563723519\_2\_full.png} & 
\texttt{N1568716881\_1\_full.png} \\
\texttt{N1569843151\_1\_full.png} &
\texttt{N1569846041\_1\_full.png} \\
\texttt{N1593401638\_1\_full.png} &
\texttt{N1597194233\_1\_full.png} \\
\texttt{N1603807807\_1\_full.png} &
\texttt{N1612279822\_1\_full.png} \\
\texttt{N1627326134\_1\_full.png} &
\texttt{N1640518562\_1\_full.png} \\
\texttt{N1644787173\_1\_full.png} &
\texttt{N1669844230\_1\_full.png} \\
\texttt{N1696193237\_1\_full.png} &
\texttt{N1699263827\_1\_full.png} \\
\texttt{N1807429484\_1\_full.png} &
\texttt{N1823505159\_1\_full.png} \\
\texttt{N1831457524\_1\_full.png} &
\texttt{N1880281788\_1\_full.png} \\ \hline \hline
    \end{tabular}
    \label{tab:cassini_iss_nac_images}
\end{table}

\section{Working with SPICE Kernels}
Each \textit{Cassini} ISS NAC image is accompanied by a \texttt{.LBL} file that houses the respective image's metadata. The \texttt{TARGET\_NAME} and \texttt{IMAGE\_TIME}  metadata entries are the necessary pieces of information for building the apparent horizon conic $\mathcal{C}$. \texttt{TARGET\_NAME} identifies the imaged planetary body, and \texttt{IMAGE\_TIME} establishes the epoch. With \texttt{TARGET\_NAME}s known, SPICE \cite{acton1996ancillary,acton2018look,Annex2020} provides its shape matrix $\mathcal{A}_{P}$ in planet-fixed coordinates $P$. With \texttt{IMAGE\_TIME} known, SPICE \cite{acton1996ancillary,acton2018look,Annex2020} provides the spacecraft's planet-relative position $r_P$ (in $P$ coordinates) and planet-relative attitude $T_P^{B}$ (i.e., coordinate transformation) from $P$ to spacecraft body-fixed coordinate system $B$. Combined, $\mathcal{A}_P$, $r_P$, and $T_{P}^B$ provide $\mathcal{C}_B$ in $B$ coordinates.

Since image processing provides $\mathcal{C}_{C}'$ in camera coordinates $C$, we require transformation $T_{B}^{C}$ to obtain $\mathcal{C}_C$ to then apply our camera calibration algorithm. Transformation
\begin{equation}
    T_{B}^{C} = T_{F}^C T_{B}^F
\end{equation}
defines the coordinate transformation from $B$ to $C$ with the focal plane coordinate $F$ as an intermediate coordinate systems. As a charge-coupled device, the \textit{Cassini} ISS NAC records measurements in $F$. SPICE reports the $T_{F}^B$ transformation as a $3-2-1$ sequence of Euler rotations with Euler angles $\theta_3 \approx 89.93^\circ$, $\theta_2 \approx -0.04^\circ$, and $\theta_1 \approx -89.99^\circ$. Once in $F$, a $180^\circ$ rotation about the camera boresight (i.e. $\hat{f}_3 || \hat{c}_3$) defines transformation $T_F^C$. Together $T_{F}^B$ and $T_F^C$ produce $T_B^C$ which ultimately provides $\mathcal{C}_C$ in desired $C$ coordinates.

\bibliographystyle{IEEEtran}
\bibliography{references}

\begin{IEEEbiography}[{\includegraphics[width=\textwidth]{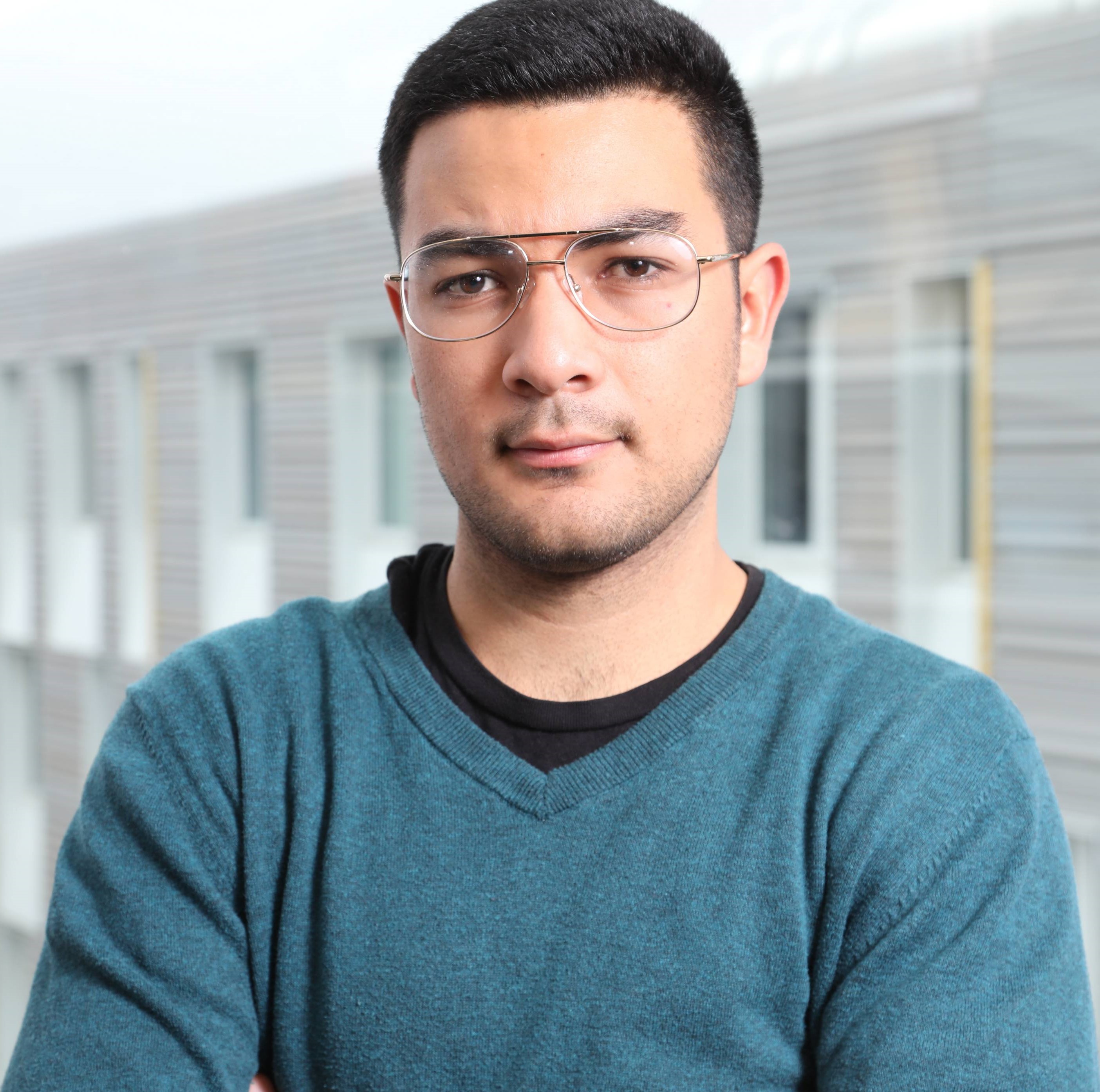}}]{Kalani R. Dana Rivera}
received his B.S. in Mechanical Engineering  at the University of Hawaii at Manoa and his M.S. and Ph.D degrees in Aerospace Engineering from Cornell University.  Kalani's research interests lie in spacecraft navigation and state estimation.
\end{IEEEbiography}

\begin{IEEEbiographynophoto}{Mason A. Peck}
received the B.S. degree in aerospace engineering from the University of Texas at Austin, Austin, TX, USA, in 1994, and the M.S. and Ph.D. degrees from the University of California, Los Angeles, Los Angeles, CA, USA, in 1999 and 2001, respectively.
From 1993 to 1994, he was with Bell Helicopter on structural dynamics. From 1994 to 2001, he was an Attitude Dynamics Specialist and Systems Engineer with Hughes Space and Communications (now Boeing Satellite Systems). From 1998 to 2001, he was a Howard Hughes Fellow. During his years at Boeing, he served as an Attitude Dynamics Lead with the Boeing Mission Control Center, participating in real-time spacecraft operations and helping to resolve spacecraft performance anomalies. In 2001, he joined Honeywell Defense and Space Systems, where he became a Principal Fellow in 2003. He has several patents on his name. In July 2004, he joined as a Faculty with Cornell University, where he teaches courses in dynamics and control and in the mechanical and aerospace engineering program, where he was promoted to an Associate Professor in Fall 2010. In 2012, he was appointed as NASA's Chief Technologist.
\end{IEEEbiographynophoto}

\end{document}